\pdfoutput=1

\documentclass[11pt]{article}

\usepackage{acl}
\usepackage{acl}
\usepackage{times}
\usepackage{latexsym}
\usepackage{multirow}

\usepackage[T1]{fontenc}

\usepackage[utf8]{inputenc}

\usepackage{microtype}
\usepackage{ulem}
\usepackage{graphicx}
\usepackage{booktabs}
\usepackage{tabularx}
\usepackage{xspace}
\usepackage{caption}
\usepackage{subcaption}
\usepackage{amsfonts}
\usepackage{amsmath}
\newcommand{\figref}[1]{Fig.~\ref{fig:#1}}
\newcommand{\tabref}[1]{Tab.~\ref{tab:#1}}
\newcommand{\secref}[1]{Sec.~\ref{sec:#1}}
\newcommand{\secrefshort}[1]{\S\ref{sec:#1}}

\newcommand{\seclabel}[1]{\label{sec:#1}}

\newcommand{\ingenuine}[0]{\texttt{ingenuine}\xspace}
\newcommand{\nongeneralizable}[0]{\texttt{ungeneralizable}\xspace}
\newcommand{\everyfeature}[0]{\texttt{every-word}\xspace}

\newcommand{\draftcomment}[3]{{\textcolor{#3}{[#1]#2}}}
\newcommand{\roy}[1]{\draftcomment{#1}{\textsc{roy}}{red}}
\newcommand{\royst}[1]{\roy{\sout{#1}}}
\newcommand{\gabi}[1]{\draftcomment{#1}{\textsc{gabi}}{blue}}
\newcommand{\gabis}[1]{\gabi{#1}}

\newcommand{\gabist}[1]{\gabi{\sout{#1}}}

\newcommand{\resolved}[1]{}
\newcommand{\com}[1]{}

\newcommand{\name}[0]{spurious correlations\xspace}
\newcommand{\uname}[0]{Spurious correlations\xspace}
\newcommand{\auname}[0]{Spurious Correlations\xspace}

\newcommand{\positivelabel}[0]{$+$\xspace}
\newcommand{\negtivelabel}[0]{$-$\xspace}

\title{On the Limitations of Dataset Balancing:\\The Lost Battle Against \auname}

\author{Roy Schwartz \qquad
  Gabriel Stanovsky \\
  School of Computer Science, The Hebrew University of Jerusalem \\
  \texttt{\{roy.schwartz1,gabriel.stanovsky\}@mail.huji.ac.il}}

\begin{document}
\maketitle
\begin{abstract}
Recent work has shown that deep learning models in NLP are highly sensitive to low-level correlations between simple features and specific output labels, leading to overfitting and lack of generalization.
To mitigate this problem, a common practice is to balance datasets by adding new instances or by filtering out ``easy'' instances~\citep{sakaguchi2020winogrande}, culminating in a recent proposal to eliminate single-word correlations altogether~\citep{Gardner:2021}.
In this opinion paper, we identify that despite these efforts, increasingly-powerful models keep exploiting ever-smaller spurious correlations, and as a result even balancing all single-word features is insufficient for mitigating all of these correlations. 
In parallel, a truly balanced dataset may be bound to ``throw the baby out with the bathwater'' and miss important signal encoding common sense and world knowledge. We highlight several alternatives to dataset balancing, focusing on enhancing datasets with richer contexts, allowing models to abstain and interact with users, and turning from large-scale fine-tuning to zero- or few-shot setups.


\end{abstract}

\section{Introduction}


Effective human communication relies on our ability to understand extra-textual context based on common sense, world knowledge or shared cultural experiences,
a property often cited as Grice's second {maxim of quantity}:~``Do not make your contribution more informative than is required''~\cite{grice1975logic,grice1989studies}.
Studies have estimated that only 12\% of the information conveyed by text is mentioned explicitly~\citep{graesser2013prose,tandon-etal-2020-dataset}.
To illustrate this, consider the question ``\textit{who is the president of the U.S.}?''. To answer it,
a human reader is likely to presume many unstated propositions, as exemplified in \tabref{example}.

\begin{figure}[tb!]
    \centering
   
    \includegraphics[width=0.49\textwidth]{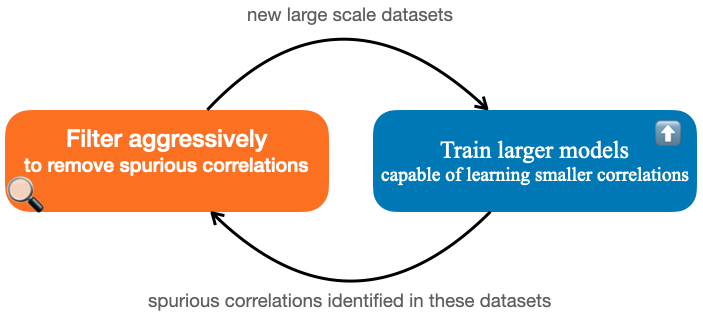}
    
    \caption{A high-level overview of the current state of supervised NLP research. Dataset developers create more aggressive filtering techniques (left), leading to larger models that are able to solve them by finding more elusive spurious correlations (right).}
    \label{fig:arms-race}
\end{figure}

\begin{table}[!t]
\centering
\begin{tabularx}{\linewidth}{@{}l X X @{}}
  \multicolumn{2}{c}{\textit{Who is the president of the U.S.?}}\\
    \toprule
{\bf Context} & {\bf Answer} \\
\midrule
\multicolumn{1}{c}{$\emptyset$} & Joe Biden \\
\textit{The year 2019} & Donald Trump\\
 \textit{The West Wing, season 1} & Josiah  ``Jed'' Bartlet \\

  \bottomrule
\end{tabularx}
\caption{\label{tab:example} Context, whether explicit or implicit, matters in textual understanding, as exemplified by the question ``\textit{who is the president of the U.S.}?''. E.g., in the first line, given no other context, a QA system should provide the most sensible fallback answer (\textit{Joe Biden}, at the time of writing).}
\end{table}

In contrast to humans, supervised models often fail to generalize and understand implicit context, instead resorting to low-level correlations in the data, leading to amplified bias~\citep{Zhao:2017,Stanovsky:2019} and brittle performance~\citep{Schwartz:2017,Gururangan:2018}\resolved{ \gabis{Do we have more recent refs? So Sam won't be angry with us ;) I tried to find ones but didn't find anything canonical}}.
To address this, recent approaches have suggested mitigating such correlations by \textit{balancing} the dataset via either adding or removing certain instances \cite{Goyal:2017,hudson2019gqa,zellers-etal-2018-swag,sakaguchi2020winogrande}. In parallel, developers keep building larger and larger pretrained models \cite{Devlin:2019,liu2019roberta,Raffel:2020}, which, when \textit{fine-tuned} on these datasets, consistently manage to reach human performance. Taken together, these trends lead to an arms-race between data curation and model development (\figref{arms-race}).
In this position paper, we question the value of mitigating \name via dataset balancing, by showing that their existence in large training sets is both inevitable and to some extent even desired, as they are an inherent property of natural language understanding. 
We build on a recent result by \citet{Gardner:2021}, who assumed that every single-word feature correlation is \textit{spurious}, i.e., can be used to mislead a model. 
We extend their argument, showing that balancing single-word features is insufficient for eliminating all \name, and that balancing feature combination is needed for that purpose. On the other hand, we show that balancing too much leads to datasets that contain no learnable signal either. We conclude by questioning whether  mitigating all \name via dataset balancing is \textit{practical}.

Following, we show that this practice is also \emph{undesired}. We show that ignoring these correlations will hinder the learning of fallback options for  both world knowledge facts (\textit{Joe Biden is the president of the U.S.}) and common sense knowledge (\textit{a person is happy when receiving a gift}), thus preventing models from using this knowledge in cases of uncertainty.
We conclude that the existence of \name in training sets should not be solved by creating more balanced datasets.\footnote{We emphasize that balancing methods are still useful as they can lead to mitigation of \textit{some} \name, and therefore better generalization \cite{bras2020adversarial,Swayamdipta:2020}, as well as potentially more efficient training. We argue that these methods are inherently limited in their ability to mitigate \textit{all} \name.}

\com{We then discuss the benefits of \name in enabling small models that are substantially more efficient, thus decreasing financial and environmental costs \cite{Strubell:2019,Schwartz:2020}.
We conclude that mitigating \name is neither practical nor desired.}

We then discuss alternatives to mitigating \name.
We argue that models should be trained to understand constructions emanating from an apriori theory of language,  such as negation, sarcasm, humor, and metaphors. 
We also suggest adopting modeling approaches that identify when the context is insufficient. We argue that in such cases, the model should \textit{not} fallback to default assumptions, but rather abstain\com{ output an ``I don't know'' response(e.g., unanswerable questions, \citealp{Rajpurkar:2018,sulem-etal-2021-know-dont})} or interact with the user to clear ambiguities.
Finally, we question the basic procedure of large-scale fine-tuning, and suggest focusing on zero- and few-shot learning instead \cite{Liu:2021}.

\section{Dataset-Model Arms Race}
\label{sec:background}

This section provides a view of recent research in NLP as an \textit{arms race} between  models and datasets. Below we describe the conditions leading to this arms race, and present our main research question, challenging its value for making progress in NLP. 

\paragraph{Models exploit \name}
While pretrained models consistently perform well across multiple tasks, various studies have pointed out that this is often achieved by exploiting \name{} in datasets, rather than improving on the underlying task~\cite{glockner_acl18,Gururangan:2018,elazar-etal-2021-back}, and that this phenomenon becomes more prominent as the models grow in size \cite{Li:2021}.

\paragraph{Mitigating \name via balancing}

\begin{figure}[tb!]
    \centering
   
    \includegraphics[width=0.3\textwidth,clip,trim={0 6.5cm 12cm 0}]{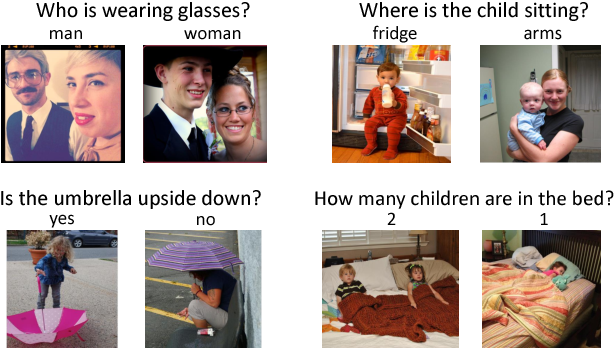}
    
    \caption{An example of dataset balancing (adapted from \citealp{Goyal:2017}). For each (question, image) pair in the VQA dataset (left), VQA2.0 adds another image, for which the answer  is different (right).
    \label{fig:vqa_balance}}
\end{figure}

Various dataset curators have tried to prevent models from learning \name by modifying  their training data via a careful control for the training label distribution, effectively striving for a \textit{balanced} dataset. 
One approach is to \textit{add} examples in order to balance the dataset~\cite{Goyal:2017,sharma-etal-2018-tackling,hudson2019gqa}. For instance, the VQA2.0 dataset \cite{Goyal:2017} is built by taking every (question $q$, image $i$, answer $a$) triplet in the VQA dataset~\citep{antol2015vqa}, and adding another triplet with the same question $q$, but a different image $i'$, guaranteed to lead to a different answer $a'$. See \figref{vqa_balance} for an example.

\paragraph{Filtering as balancing}
A complementary balancing approach to augmentation is \textit{filtering} examples out from datasets such that \name are minimized. This approach was taken in the creation of the  SWAG dataset \cite{zellers-etal-2018-swag}, using ``adversarial filtering'' (AF). In AF, dataset instances that are easily solved by an adversarial model are filtered out.
The AF approach and similar approaches were picked up by many datasets such as ReCoRD \cite{Zhang:2018}, DROP~\citep{dua-etal-2019-drop}, HellaSWAG~\cite{zellers-etal-2019-hellaswag}, $\alpha$NLI \cite{Bhagavatula:2020}, and WinoGrande~\cite{sakaguchi2020winogrande}.

Here we argue that approaches like AF converge to removing all low-level correlations,\footnote{Indeed, AFLite, an extension of AF, was designed to ``systematically discover and filter \textit{any} dataset artifact in crowdsourced commonsense problems'' (\citealp{bras2020adversarial}, emphasis in the original).} and therefore a fully balanced dataset.
As this approach relies on an external model, applying it with ever stronger models with higher capacity, will allow these models to pick up on subtler correlations \cite{Li:2021}. At the extreme, the remaining instances that could not be solved by a fully capable model will have no statistical signal that can be exploited by that model, i.e., a balanced dataset\resolved{\gabist{ We therefore refer to both explicit balancing and filtering as balancing methods in the rest of the paper.} \roy{gabi, plz check. I wonder whether we need a more mathy argument here} \gabis{I think it looks good. I don't think we need mathy arguments. I'm not sure if we need the last sentence here which I proposed to remove.}}.\resolved{\roy{maybe add something?}} We henceforth refer to both augmentation and filtering as \textit{balancing} methods.

\paragraph{Large models solve the new datasets}
In parallel to the efforts in dataset balancing, the leading \textit{modeling} approach in recent years in NLP is pretraining large language models on raw text corpora, followed by fine-tuning them on supervised downstream applications.
These models continue to grow in size~\citep{Peters:2018,Devlin:2019,liu2019roberta,radford2019language,Raffel:2020}, and their fine-tuning performance improves accordingly.
This in turn leads to more aggressive balancing, setting in motion a kind of \textit{arms race} between datasets and models (\figref{arms-race}). 

Evidently, a similar trend emerges for the  previously mentioned datasets: 
(1) the first baselines, reflecting the state of the art at the time of dataset creation, perform relatively poorly, e.g., 59\% on SWAG, 47\% on ReCoRD, 47 F1 on DROP, 47\% on HellaSWAG, 69\% on $\alpha$NLI, and 79\% on WinoGrande; 
(2) model developers introduce increasingly larger and heavily-parameterized models, hill-climbing on these datasets; and eventually
(3) models essentially solve the dataset within a year or two, often outperforming humans: 86\% on SWAG~\citep{Devlin:2019}, 94\% on ReCoRD \cite{He:2021a}, 88 F1 on DROP~\citep{chen-etal-2020-question}, 93\% on HellaSWAG \cite{He:2021a}, 92\% on $\alpha$NLI~\citep{He:2021b}, and 90\% on WinoGrande~\citep{Raffel:2020}.
(4) new large-scale datasets are collected with more aggressive pruning techniques, thus repeating the cycle.

Based on these findings, our main research question is whether dataset balancing is the most promising method for mitigating \name.
We note that an arms race between models and  datasets might spur advances. Here we question a specific aspect of this arms race: the improvement of datasets by using more aggressive filtering techniques.
Next we turn to present practical and conceptual limitations of this practice.



\section{The Lost Battle Against \auname}
\label{sec:lost-battle}
\begin{table}[!t]

\setlength{\tabcolsep}{3pt}
\small
\centering
\begin{tabularx}{\linewidth}{l X c} 
  \toprule
\textbf{Name}&  {\bf Description} \\
\midrule
{\ingenuine}& Correlations between features and output labels for no reason.\\
 \nongeneralizable & Correlations that do not generalize to new contexts.\\
 \everyfeature & Correlations between every single-word feature and output label. \\
\bottomrule
\end{tabularx}
\caption{\label{tab:spurious} Different definitions of \textit{\name}.}
\end{table}

So far we have identified dataset balancing as  a common way to mitigate \name{}. Next, we outline how different works define \name{}\resolved{\royst{, and adopt a working definition for this paper, which treats every feature combination as potentially spurious}\roy{We no longer take that strong position}}~(\secref{name:def}), and then question whether dataset balancing is a viable way for mitigating them; 
we note that balancing too little is bound to leave \name{} in the data~(\secref{balance:too-little}), while balancing too much 
discards meaningful signal~(\secref{balance:too-much}). We finish by questioning whether this practice is even desired (\secref{undesired}).

\subsection{What are \auname?}
\seclabel{name:def}

Mitigating \name is frequently used as motivation for developing new balancing approaches. However, the term \textit{\name} is often not clearly and consistently defined. The basic definition is a set of features that are correlated but not causally related.\footnote{\url{https://en.wikipedia.org/wiki/Spurious_relationship}} 

In NLP, several definitions of \name are typically used. One conceptual definition, denoted here \ingenuine (e.g., \citealp{wang-culotta-2020-identifying,rogers-2021-changing}) is a feature correlated with some output label for no apparent reason. Such features often result from the annotation process (referred to as \textit{annotation artifacts}; \citealp{Gururangan:2018}). For instance, \citet{Gururangan:2018} have shown that the words ``cat'' and ``sleeping'' are  correlated with contradictions in the SNLI dataset \cite{bowman2015large}. 

This definition is appealing: we want our models to learn real information about the world, and not properties of a given dataset. However, it is also somewhat subjective, and could include features that might be referred to as genuine, such as the word ``not'' indicating NLI contradictions. Further, genuine features, i.e., those representing a real phenomenon in the world (e.g., ``amazing'' as a feature for positive sentiment), are also likely to lead models make to erroneous predictions in some contexts (e.g., negation or sarcasm; \citealp{Gardner:2021}). Such features could thus harm generalization, so some might consider them spurious as well.\footnote{See \citet{Eisenstein:2022} for discussion of different feature types.}

In an alternative definition, denoted \nongeneralizable, a spurious feature is one that 
works well for specific examples but does not hold in general 
\cite{chang-etal-2021-robustness,yaghoobzadeh-etal-2021-increasing}. This definition does not address the nature of the feature (genuine or not), but does make an implicit assumption that such features are of high importance (e.g., high pointwise mutual information values with the corresponding label; \citealp{Gururangan:2018}). This definition is no longer subjective in terms of the genuineness of the feature, but is still subjective in the level of effect on generalizability (i.e., what is a \textit{high} value of  PMI?). 

\newcite{Gardner:2021} relaxed the last constraint, and assumed that \textit{every} simple correlation between single word features and output labels is spurious (henceforth \everyfeature). 
They then defined a class of \emph{competent} datasets, where the marginal probability for every feature is uniform over the class label, i.e., for any feature $x_i$ and label $y \in Y$, $p(y|x_i) = \frac{1}{|Y|}$, thus limiting models from picking up any correlation between single features and output labels. 

We next extend the \everyfeature approach beyond single words, showing that models that can exploit single word features can also exploit some feature interactions, and therefore these should also be considered spurious. 
\tabref{spurious} summarizes the different definitions of \name.


\subsection{Balancing too Little Leaves some Spurious Features}
\seclabel{balance:too-little}
\citet{Gardner:2021} assumed that as each word can appear in certain contexts that change its semantic meaning (e.g., negation, sarcasm), each word is potentially spurious. 
Here we note that the same argument can be applied to 
feature interactions, such as word $n$-grams. 
We start with a toy example to illustrate our argument for bigrams, and then extend it for larger values of $n$. 

\begin{table}[!t]

\setlength{\tabcolsep}{5pt}
\small
\centering
\begin{tabularx}{.6\linewidth}{c X c} 
  \toprule
\textbf{Split}&  {\bf Text} & {\bf Label} \\
\midrule
 \multirow{4}{*}{\textit{Train}}& very good & \positivelabel\\
 & very bad & \negtivelabel\\
 & not good & \negtivelabel\\
 & not bad & \positivelabel\\
\midrule
  \multirow{2}{*}{\textit{Test}}& not very good & \negtivelabel\\
 & good & \positivelabel\\
\bottomrule
\end{tabularx}
\caption{\label{tab:balanced_example} A toy example of a  training set (\textit{Train}), which is balanced for unigrams, but not for bigrams. Relying on the bigram correlations (e.g., memorizing that ``very good'' leads to a positive sentiment) will lead to mispredictions on the test set (\textit{Test}).}
\end{table}


Consider the toy dataset for the task of sentiment analysis shown in \tabref{balanced_example}, with vocabulary $V$=\{\textit{good, bad, not, very}\}, and label set $Y$~=~\{~\positivelabel, \negtivelabel~\}. 
The \textit{Train} split is balanced with respect to single-word features, i.e., it is a \textit{balanced} or \textit{competent} dataset:
\[
\forall{w \in V, y \in Y}~:~p(y|w)=\frac{1}{|Y|} 
\] 
 
Assume the semantics of this dataset is that of  English\com{ (e.g., \textit{not} in this dataset means \textit{not} in English)}, while `\positivelabel' means positive sentiment and `\negtivelabel' means negative. 

A model trained on \textit{Train} can achieve perfect training accuracy by learning the correct semantics. However, achieving perfect training accuracy can also be done by learning correlations between \textit{two-word} features and the target label (i.e., memorizing all the training examples). In this case, the model would make the wrong prediction for the first test example in \textit{Test} (as it has learned that \textit{very good} 
 is a feature that indicates positive sentiment), and similarly, will make a random prediction for the second test example, which does not contain any two-word feature seen during training. 

This example highlights that balancing single-word features does not guarantee resiliency to \name, and therefore in order to mitigate all \name, balancing pairs of features is also required. 
One can construct similar examples for larger values of  $n$, by similarly considering multi-word expressions and common co-occurrences (e.g., ``jaw dropping'', ``worst day ever''). These could serve as \name in the same way single words do.

Another example is sarcasm. A model that fails to understand sarcastic contexts will misinterpret statements that appear in such contexts, even if it perfectly understands the base meaning of these statements. Thus, the entire reasoning process of such a model, whether relying on simple features, feature interactions, or other types of understanding, will result in mispredictions of certain inputs, and thus can be considered spurious. 

As a result, to truly mitigate all \name in a dataset, balancing feature combinations is required as well. Accordingly, balancing too little will leave some \name in the dataset.

\subsection{Too much Balancing Prevents Learning Valuable Semantic Knowledge}
\seclabel{balance:too-much}

We observed that balancing too little does not allow models to fully eliminate \name. Here we show that too much balancing can prevent models from learning valuable knowledge. 

\begin{table}[!t]

\setlength{\tabcolsep}{5pt}
\small
\centering
\begin{tabularx}{.72\linewidth}{c c | c c} 
  \toprule
\multicolumn{2}{c}{Original Train Set} & \multicolumn{2}{c}{Augmented Samples} \\
{\bf Input} & {\bf Label} & {\bf Input} & {\bf Label} \\
\midrule
 0 0 & 0 & *0 0 & 1\\
 0 1 & 1 &*0 1 & 0\\
 1 0 & 1&*1 0 &0\\
 1 1 & 0&*1 1 &1\\
\bottomrule
\end{tabularx}
\caption{\label{tab:xor} Left: a training set for the XOR function, balanced for unigrams. Right: requiring that bigrams are also balanced would prevent models from learning.}
\end{table}

Consider the training data for learning the XOR function presented in \tabref{xor} (left). 
This dataset contains enough learnable signal when considering feature interactions despite being balanced for single words. Nonetheless, balancing this dataset for \textit{pairs of features} would result in no information, and thus prevent any model from learning this function (\tabref{xor}, right). 

Now consider a given natural language dataset $D$. Define $n$ to be the length of the longest document in $D$. By definition, balancing every combination of up to $n$ features leaves no learnable signal in $D$.\footnote{We assume the standard data collection process when using AF, in which the last step is balancing \cite{zellers-etal-2018-swag,dua-etal-2019-drop}, and longer instances cannot be added.} We conclude that balancing too much can prevent models from learning semantic knowledge.

Combining the two observations, we are left with the question of the potential intersection between balancing too much and balancing too little: does a sweet spot exist for which no \name are found in the dataset, but enough learnable signal is left?
And even if so, would a balancing algorithm, whether by augmentation or filtering, be able to find it? 
We leave these questions for future work, but note that addressing them is a prerequisite for the theoretical and practical application of dataset balancing  for mitigating \name{}.

\subsection{Dataset Balancing is \textit{Undesired}}
\label{sec:undesired}
Even if a sweet spot exists between balancing too little and too much, do we really want to find it? Here we argue that perhaps not.

The practice of dataset balancing is designed to prevent models from learning that some words or expressions have a common fallback meaning that can stem from dataset artifacts (e.g., ``cat'' as an indicator of contradiction) but also from cultural and historical contexts (e.g., Biden is the  U.S.~president in 2022).
Fallback meanings are crucial for understanding language, as contexts are often underspecified~\cite{graesser2013prose}. 
Indeed, relying on fallback meanings might make models fail to process some inputs correctly, and might not generalize to other domains where the fallback meaning is different. We argue that the ability to use them is a central ability of language understanding. 

\com{
Simple examples are adjectives like ``amazing'', ``great'' and ``terrbile''. A sentiment analysis model that observes each of these words with equal probability for both positive and negative sentiments, might make a wrong prediction for single-word reviews, e.g., the single-word review ``great''. 

But this problem goes beyond simple adjectives. }For example, substantial efforts are made to teach models \textit{world knowledge}, such as that the president of the U.S.~is Joe Biden, the capital of Brazil is Brasília, and France is the soccer world champion. These efforts include building world knowledge datasets~\cite{Wang:2021}, developing methods for enhancing models with this information \cite{zhang-etal-2019-ernie,Peters:2019}, and evaluating how well models capture it \cite{Rubinstein:2015,roberts2020knowledge}.
But many of these world-knowledge facts are context dependent: the capital of the Brazil has changed in 1960, the president of the U.S., as well as soccer world champions potentially change every 4 years, etc. 

Another example is \textit{common sense knowledge}, such as ``people are happy when they receive a gift'', ``an elephant is taller than a zebra'', and ``a statue that doesn't fit into a suitcase is too large''. 
A large body of work has been carried out to create benchmarks that measure the common sense abilities of models \cite{Liu2004,Levesque:2012,zellers-etal-2018-swag,sakaguchi2020winogrande,Bisk:2020}, as well as augmenting models with such abilities \cite{Qin:2020,Bosselut:2021}. 

Common sense reasoning is, by definition, stochastic and reliant on understanding presupposed, underspecified context. One could imagine a person unhappy to receive a gift (e.g., because it is not what they wanted), a fantastically large zebra compared to a tiny elephant, and a suitcase with multiple compartments which prevent a small statue from fitting in it.

These examples illustrate that a model that learns these correlations and relies exclusively on them to make predictions is limited and is bound to make mistakes in some contexts. 
One way to avoid these mistakes is to balance these correlations out, and prevent models from knowing these assertions to begin with. We argue that this solution is not a \textit{desired} solution. In essence, an interpreter's task (be it human or machine) is to infer the most probable context in which a statement is made, and as a result, it \textit{should} have a fallback option for such world knowledge and common sense assertions.


\paragraph{Discussion}
We recognize that a balanced dataset may not be balanced with respect to the appearance of common-sense or world-knowledge assertions \textit{in a given context}. E.g., a model might balance-out the general fact that Joe Biden is the U.S.~president, but \textit{not} that he is the president in 2022. As in many cases much of the context is unobserved \cite{graesser2013prose}, the question is whether we want models to make a prediction in cases of uncertainty based on the fallback option. We argue that doing so is a desired strategy in many cases (though a preferred strategy might be to interact of abstain from making a decisive prediction, see \secref{abstention}).

We also acknowledge that correlations in the real world can be misleading. For instance, people often mistake the biggest commercial city in some countries for their capital (e.g., Istanbul in Turkey), potentially due to the high correlation between the two. In such cases, relying on the fallback option might lead to prediction error. 
However, we argue that following the human strategy of relying on a fallback option in cases of uncertainty will promote models' communication abilities.
\footnote{A counter example is social biases, where we want to explicitly discourage models from having a fallback option (see \secref{social-bias} for discussion).}

We want to stress that balancing methods can result in mitigating \textit{some} of the \name, and therefore lead to increased generalization~\cite{bras2020adversarial,Swayamdipta:2020}. 
Moreover, the process of filtering the data naturally results in smaller datasets, which leads to lower training costs~\cite{Swayamdipta:2020}. While such contribution is meaningful in terms of, e.g., environmental concerns~\citep{Strubell:2019,Schwartz:2020}, it is orthogonal to our research question.
Overall, despite the important contributions of balancing techniques, this paper shows that even the perfect balancing method might not mitigate all \name in a satisfying way.


So how can we make models more resilient to \name without balancing the data? Below we discuss several ideas for doing this.

\section{Ways to Move Forward}
\label{sec:ways-forward}

So far, we presented limitations of dataset balancing as a means to mitigate \name. In this section 
we discuss several alternatives to this practice, summarized in \tabref{ways-forward}. 
We note that none of these proposals is particularly novel. 
Rather, we intend to survey alternatives proposed in literature and argue
that these may be promising for addressing the drawbacks of \name, and that more efforts should be put into studying them.

\begin{table}[!t]
 \setlength{\tabcolsep}{3pt}
\centering
\begin{tabularx}{\linewidth}{@{}l X X @{}}
    \toprule
{\bf Current Practice} & {\bf Proposal} \\
  \midrule
 {Dataset balancing} & Richer contexts (\secrefshort{richer})\\
A closed label set & Abstain/interact (\secrefshort{abstention})\\
Large-scale fine-tuning & Few-shot learning (\secrefshort{end-ft})\\
   \bottomrule
\end{tabularx}
\caption{\label{tab:ways-forward} Our suggestions for mitigating the effects of \name, listing three current practices, each with an alternative proposal.}
\end{table}

\subsection{Augmenting Datasets with Rich Contexts}\label{sec:richer}

The implicit assumption of dataset balancing is that in order to mitigate \name the model has to \textit{unlearn} them, that is, they should be removed altogether from the training set. 
We argue that instead we should be focusing on learning and modeling richer contexts. 


As an example, consider negation. A model that generalizes well, should learn the meaning of words such as \textit{not}, and should be able to negate new words, even those that were seen only in positive contexts at training time. For example, if a model only sees during training  words like ``amazing'' or ``happy'' with positive sentiment, and thus learns that these words bear positive meaning, we would still expect it to interpret their negated appearance (e.g., \textit{not  amazing}) as an indicator of \textit{negative} sentiment. Such generalization is crucial for language learning, and should ideally allow models to not rely exclusively on \name. Despite the immense progress in the field in the past decade, negation still poses a challenge to modern NLP tools~\cite{hossain-etal-2020-non,Hossain:2022}.\footnote{Though we should continually assess the challenge negation poses on the most recent models \cite{Bowman:2021}.}

We suggest taking into account different types of contexts during dataset design. In particular, collecting training examples with contexts such as negation \cite{morante-blanco-2012-sem}, humor \cite{weller2019humor,Annamoradnejad:2021}, sarcasm \cite{davidov-etal-2010-semi,Oprea:2020}, or metaphors \cite{tsvetkov-etal-2014-metaphor,mohammad-etal-2016-metaphor}. This recommendation applies to both supervised tasks, and perhaps more so to pretrained data.
We suggest adding documents with such contexts throughout the pretraining corpus, or as a continued pretraining step to existing large-scale models.\footnote{We recognize that editing pretrained corpora poses significant challenges due to their immense size, as demonstrated by recent efforts such as corpus analysis \cite{dodge2021documenting} and deduplication \cite{lee2021deduplicating}.}

To incorporate contexts from a wide range of phenomena, we can leverage the vast literature on broad-coverage semantics~\citep{framenet,ccg,amr,ucca}.\footnote{See \citet{abend-rappoport-2017-state} for a survey.} This line of work proposes  theories of language, composing inventories of linguistic constructions with an algebraic formulation of their inter-relations  in terms of truth value, factuality, and more. These inventories often include the phenomena discussed above, such as negation, sarcasm, and presupposition.

\subsection{Interaction and Abstention to Cope with Underspecified Contexts}
\label{sec:abstention}

\begin{figure}[tb!]
    \centering

    \includegraphics[width=0.48\textwidth]{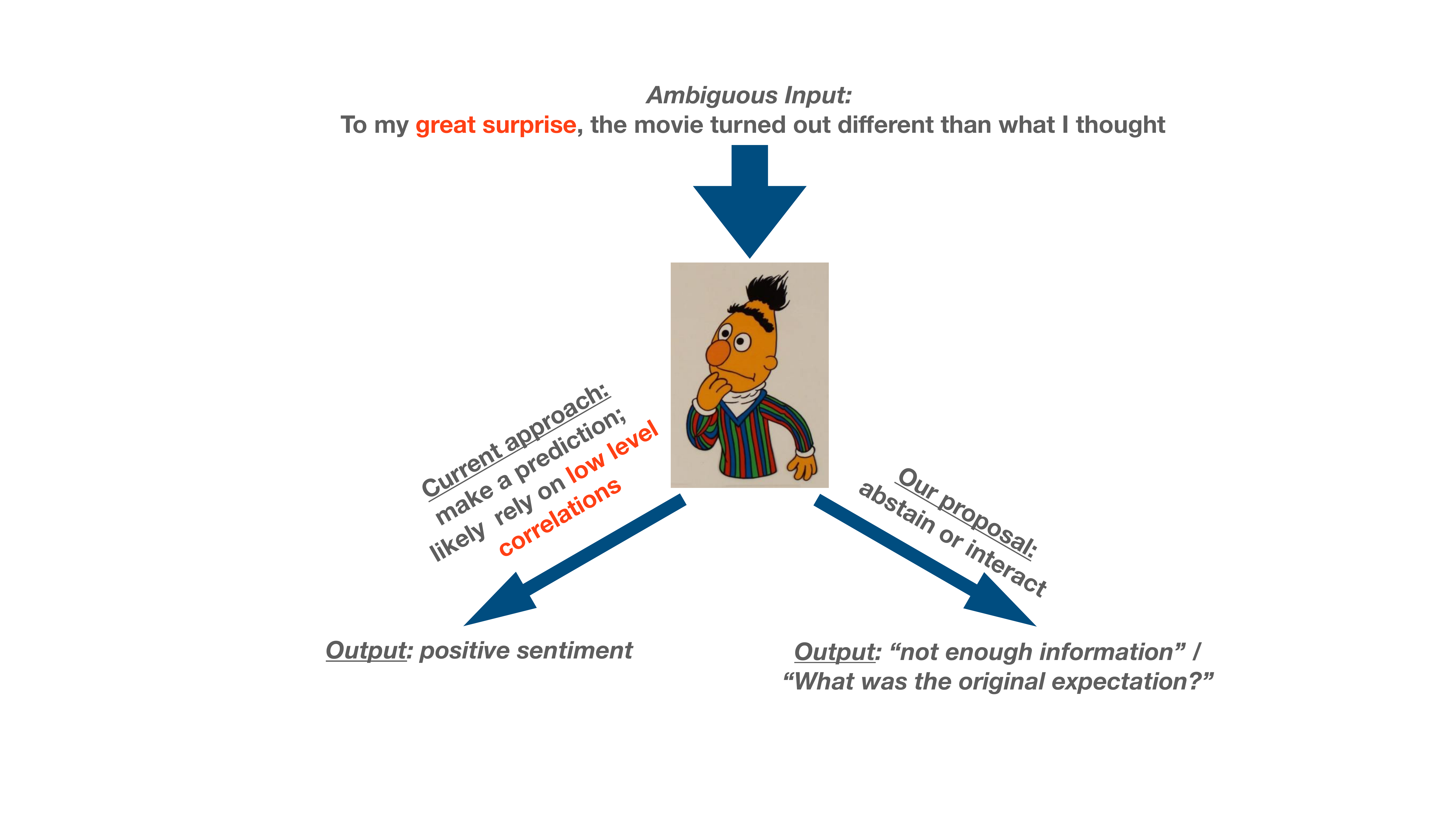}
    
    \caption{An example of abstention/interaction in cases of uncertainty. For the task of sentiment analysis, models currently assign a label to each input, even for ambiguous or underspecified ones (top). This may lead the model to over-rely on \name (marked in red, bottom left). Models that abstain or interact (bottom right) might learn to rely less on such correlations.}
    \label{fig:idk}
\end{figure}


Most NLP tasks are designed with a closed label set that forces models to make a concrete prediction for each test instance, without an option to abstain or interact with the user to get more information. Even for tasks with a large label set (e.g., language modeling), models still have to output a valid vocabulary item.
Here we argue that this practice creates an \textit{inductive bias} towards using \name{} in cases of uncertainty, as the model has ``nothing to lose'' in case of low certainty, and is encouraged to always make some prediction, potentially relying on \name.\footnote{We recognize that in some cases we do want the model to make a prediction under cases of uncertainty (see \secref{undesired}). The ability to detect when is it reasonable to make an educated guess is an important property of an intelligent agent, and an exciting research question.}



To further illustrate this point, consider the ambiguous sentence ``\textit{To my great surprise, the movie turned out different than what I thought}.'', in the context of sentiment analysis. The reader cannot infer whether the writer is pleasantly surprised (a positive review) or disappointed (a negative review). We argue that in such cases models might lean towards a positive sentiment based on the words ``great'' and ``surprise'', which are typically correlated with a positive sentiment. 

To test this, we ran a RoBERTa-large model \cite{liu2019roberta} fine-tuned on SST-2 \cite{socher-etal-2013-recursive} on that example.\footnote{We used the AllenNLP demo (\url{https://demo.allennlp.org/sentiment-analysis/}).} As expected, the model returns a \textit{positive} label, with 99.99\% confidence. Interestingly, three different interpretation methods (simple gradient visualization, \citealp{simonyan2014deep}; integrated gradient visualization, \citealp{sundararajan2017axiomatic}; and SmoothGrad, \citealp{smilkov2017smoothgrad}) all find the word ``great'' or ``surprise'' to be one of the three most influential words on the model's prediction. While this example does not prove the prevalence of this problem, it does demonstrate its existence.

To address this problem, we suggest adopting approaches that allow models to abstain and interact when they cannot make a decision with high confidence \cite{Chow1957AnOC, Hellman1970TheNN, Laidlaw:2019,balcan2020power}. See \figref{idk}. 
This can be achieved by building datasets with unanswerable questions \cite{Ray:2016,Rajpurkar:2018,
sulem-etal-2021-know-dont}, but also by designing models that abstain in cases of low certainty for all inputs,  even those with an unambiguous gold label.\footnote{Model calibration techniques  \cite{Degroot:1983,Guo:2017,card-smith-2018-importance} are often used  both for allowing models to abstain \cite{Cortes:2016,shrikumar2019flexible} and identifying unanswerable questions \cite{kamath-etal-2020-selective,zhang-etal-2021-knowing}.} 
We hypothesize that encouraging the model to provide this output when it is unsure, rather than making a semi-educated guess, potentially based on \name, could reduce its dependency on such correlations.

\com{\subsection{Interaction}

\subsection{Model Solutions}

\subsection{Dataset Solutions}
}

\subsection{The End of Large-Scale Fine-Tuning?}\label{sec:end-ft}


This paper has demonstrated the limitations of mitigating \name via dataset balancing. A naive way to mitigate \name is to stop using large-scale datasets altogether. We echo recent calls \cite{Liu:2021} and argue that for supervised learning (i.e., large-scale fine-tuning), recent advances in zero- and few-shot learning might make this option possible.


Large pretrained models such as T5 \cite{Raffel:2020}
or GPT-3 \cite{Brown:2020}, trained on vast amounts of data, arguably learn enough about the world to acquire many of the skills currently learned through supervised learning. Indeed, the large increase in the size and capacity of pretrained language models has led to a new wave of few-shot and zero-shot methods~\cite{schick2021its,shin2020autoprompt,gu2021ppt}, which are able to reach human-level performance on certain tasks using only a few dozens of training examples \cite{Schick:2021}.\resolved{\gabis{will this draw fire? Maybe it's better to specify exactly what skills reach human-level performance?}\roy{Better? from their abstract: \textit{PET achieves a new state of the art on RAFT and performs close to non-expert humans for 7 out of 11 tasks}.}} Given these impressive results, it is not clear whether there is still value in fine-tuning models on large-scale datasets for all tasks.
In the context of this work, focusing on few-shot learning might allow models to not learn some of the correlations that result from manual annotation \cite{Schwartz:2017,Gururangan:2018,poliak-etal-2018-hypothesis}, as they will not be exposed to many of them to begin with. 

We note that this proposal is not a perfect solution. First, some \name may be picked up by the small number of examples. This is less of a problem in the zero-shot setting, or in cases where the model parameters are not updated in few-shot settings \cite{Brown:2020}, but studying the extent to which \name are picked up in other  few-shot settings is an important avenue for future research. Second, some \name might be picked up during the pretraining stage \cite{gehman2020realtoxicityprompts,birhane2021multimodal,dodge2021documenting}. Continuing to quantify this phenomenon and finding ways to mitigate it is another important line of research.


An important question in this context is the tasks for which supervised learning is still needed. 
It seems plausible that excelling in language modeling tasks requires mastering the skills that stand in the base of many NLP tasks, such as sentiment analysis, syntactic parsing, and NER. However, it is similarly plausible that this is not the case for other tasks, e.g., summarization, simplification and dialogue.
We are cautious in making concrete recommendations for which tasks to apply this principle, but suggest the following intuitive rule of thumb: for datasets or tasks for which the state of the art is close to or surpasses the human baseline, we should consider moving to few-shot setups.


Finally, dataset creation is still a valuable and important line of research. Our recommendation to stop building large scale training sets does not make this task redundant, to both spur the design of better models, and to better test their capabilities. We suggest that instead of building large training sets and small validation and test sets, authors should consider building large test sets, as a means for achieving improved statistical power~\cite{Card:2020}.


\subsection{A Note on Social-Bias Correlations}
\label{sec:social-bias}
So far, we discussed the problems with unlearning \name, and advocated instead for more elaborate context modeling.
One exception might be the case of social biases. Textual data often reflects human stereotypes such as \name{} between labels and protected group attributes, e.g., alignments between professions and gender or race.
Unlike other types of knowledge discussed in \secref{undesired}, in this case there is an incentive to prevent models from learning this type of correlation as means for actively reducing the harms of such biases, especially in commercial and public-facing applications, such as machine translation~\citep{Stanovsky:2019} or automated financial decision-making~\citep{bartlett2021consumer}. As a result, methods for dataset balancing are no longer \textit{undesired} for mitigating such \name. 
\resolved{\gabis{I'm not sure I understand what we're saying from here until the end of the section. Should we elaborate how the methods we propose assist in mitigating societal bias?}\roy{better?}}

Nonetheless, as demonstrated in \secref{lost-battle}, methods for dataset balancing are a limited solution for mitigating \name, including social-bias ones. 
In contrast, the methods proposed in this section for mitigating \name might also assist in mitigating social biases, or at least slow down their amplification~\cite{Zhao:2017}.

\section{Related Work}

This paper discusses the arms-race between models and datasets.
Previous works criticized one side of this arms race---the increasing size of pretrained models---due to ethical and environmental concerns~\citep{Schwartz:2020,10.1145/3442188.3445922}, or questioning its ability to learn meaningful abstractions from raw text~\cite{bender-koller-2020-climbing,Merrill:2021}. 
This work studies the second part of this arms race, regarding the efforts to mitigate \name through dataset balancing. The release of such datasets is often motivated by their potential to spur progress in modeling, and to help tease apart qualitative differences between models. \citet{liu-lee-jia-liang:2021:arxiv} showed that this is not necessarily the case, by observing that the ranking of reading comprehension models on  small and synthetic benchmarks is similar to that of the (large) SQuAD dataset \cite{rajpurkar2016squad}. 

\citet{raji2021ai} recently criticized the concept of benchmarks as a whole, arguing that they are only capturing specific skills and not ``general'' capabilities. Our paper raises related concerns about training sets implicitly containing \name, and suggests reconsidering the practice of building large-scale training sets.

Finally, concurrent to this work, \citet{Eisenstein:2022} discussed several types of \name in the context of causality theory \cite{pearl2009causality}, and used a toy example to demonstrate their different effects on models. They concluded that domain knowledge is required to identify the correlations that are indeed spurious, i.e., those that might challenge the generalization ability of models.

\section{Conclusion}
\uname in large textual corpora can result in model brittleness, lack of generalization, and an inflated sense of the state of the art. Mitigating their negative side-effects is an important research goal of the NLP community. In this paper we presented practical and conceptual limitations of  dataset balancing as a means for doing so. We proposed alternative ways for mitigating \name, including adding richer contexts to textual corpora, and allowing models to abstain or interact in cases of uncertainty. We concluded by suggesting to reconsider the practice of fine-tuning pretrained models on large-scale training sets. 

 \section{Broader Impact and Ethical Consideration}
 Our work did not involve any new data or annotation collection, and as such did not require crowdsourced or in-house workers, or introduces any new models and related risks.
 Instead, we examine existing resources and common data balancing approaches.
 In Section~\ref{sec:social-bias} we specifically discuss the relation between these practices and implications on social bias in models.

\section*{Acknowledgements}
We would like to thank Matt Gardner and Will Merrill for the in-depth discussion. We would also like to thank Omri Abend, Yoav Goldberg, Inbal Magar, and the anonymous reviewers for their feedback. This work was supported in part by research gifts from the Allen
Institute for AI.\resolved{\roy{Gabi, anything else?}}

\bibliography{custom,anthology}

\begin{thebibliography}{103}
\expandafter\ifx\csname natexlab\endcsname\relax\def\natexlab#1{#1}\fi

\bibitem[{Abend and Rappoport(2013)}]{ucca}
Omri Abend and Ari Rappoport. 2013.
\newblock \href {https://aclanthology.org/P13-1023} {{U}niversal {C}onceptual
  {C}ognitive {A}nnotation ({UCCA})}.
\newblock In \emph{Proc. of ACL}.

\bibitem[{Abend and Rappoport(2017)}]{abend-rappoport-2017-state}
Omri Abend and Ari Rappoport. 2017.
\newblock \href {https://doi.org/10.18653/v1/P17-1008} {The state of the art in
  semantic representation}.
\newblock In \emph{Proc. of ACL}.

\bibitem[{Annamoradnejad and Zoghi(2020)}]{Annamoradnejad:2021}
Issa Annamoradnejad and Gohar Zoghi. 2020.
\newblock \href {https://arxiv.org/abs/2004.12765} {{ColBERT}: Using bert
  sentence embedding for humor detection}.
\newblock {arXiv}:2004.12765.

\bibitem[{Antol et~al.(2015)Antol, Agrawal, Lu, Mitchell, Batra, Zitnick, and
  Parikh}]{antol2015vqa}
Stanislaw Antol, Aishwarya Agrawal, Jiasen Lu, Margaret Mitchell, Dhruv Batra,
  C.~Lawrence Zitnick, and Devi Parikh. 2015.
\newblock \href {https://doi.org/10.1109/ICCV.2015.279} {{VQA:} visual question
  answering}.
\newblock In \emph{Proc. of ICCV}.

\bibitem[{Baker et~al.(1998)Baker, Fillmore, and Lowe}]{framenet}
Collin~F. Baker, Charles~J. Fillmore, and John~B. Lowe. 1998.
\newblock \href {https://doi.org/10.3115/980845.980860} {The {B}erkeley
  {F}rame{N}et project}.
\newblock In \emph{Proc. of ACL}.

\bibitem[{Balcan et~al.(2020)Balcan, Blum, Sharma, and Zhang}]{balcan2020power}
Maria-Florina Balcan, Avrim Blum, Dravyansh Sharma, and Hongyang Zhang. 2020.
\newblock On the power of abstention and data-driven decision making for
  adversarial robustness.
\newblock {arXiv}:2010.06154.

\bibitem[{Banarescu et~al.(2013)Banarescu, Bonial, Cai, Georgescu, Griffitt,
  Hermjakob, Knight, Koehn, Palmer, and Schneider}]{amr}
Laura Banarescu, Claire Bonial, Shu Cai, Madalina Georgescu, Kira Griffitt, Ulf
  Hermjakob, Kevin Knight, Philipp Koehn, Martha Palmer, and Nathan Schneider.
  2013.
\newblock \href {https://aclanthology.org/W13-2322} {{A}bstract {M}eaning
  {R}epresentation for sembanking}.
\newblock In \emph{Proc. of LAW VII \& ID}.

\bibitem[{Bartlett et~al.(2021)Bartlett, Morse, Stanton, and
  Wallace}]{bartlett2021consumer}
Robert Bartlett, Adair Morse, Richard Stanton, and Nancy Wallace. 2021.
\newblock Consumer-lending discrimination in the fintech era.
\newblock \emph{Journal of Financial Economics}.

\bibitem[{Bender et~al.(2021)Bender, Gebru, McMillan-Major, and
  Shmitchell}]{10.1145/3442188.3445922}
Emily~M. Bender, Timnit Gebru, Angelina McMillan-Major, and Shmargaret
  Shmitchell. 2021.
\newblock \href {https://doi.org/10.1145/3442188.3445922} {On the dangers of
  stochastic parrots: Can language models be too big?}
\newblock In \emph{Proc. of FAccT}.

\bibitem[{Bender and Koller(2020)}]{bender-koller-2020-climbing}
Emily~M. Bender and Alexander Koller. 2020.
\newblock \href {https://doi.org/10.18653/v1/2020.acl-main.463} {Climbing
  towards {NLU}: {On} meaning, form, and understanding in the age of data}.
\newblock In \emph{Proc. of ACL}.

\bibitem[{Bhagavatula et~al.(2020)Bhagavatula, Bras, Malaviya, Sakaguchi,
  Holtzman, Rashkin, Downey, Yih, and Choi}]{Bhagavatula:2020}
Chandra Bhagavatula, Ronan~Le Bras, Chaitanya Malaviya, Keisuke Sakaguchi, Ari
  Holtzman, Hannah Rashkin, Doug Downey, Scott Wen-tau Yih, and Yejin Choi.
  2020.
\newblock \href {https://doi.org/10.48550/ARXIV.1908.05739} {Abductive
  commonsense reasoning}.
\newblock In \emph{Proc. of ICLR}.

\bibitem[{Birhane et~al.(2021)Birhane, Prabhu, and
  Kahembwe}]{birhane2021multimodal}
Abeba Birhane, Vinay~Uday Prabhu, and Emmanuel Kahembwe. 2021.
\newblock \href {https://arxiv.org/abs/2110.01963} {Multimodal datasets:
  misogyny, pornography, and malignant stereotypes}.
\newblock {arXiv}:2110.01963.

\bibitem[{Bisk et~al.(2020)Bisk, Zellers, {Le Bras}, Gao, and Choi}]{Bisk:2020}
Yonatan Bisk, Rowan Zellers, Ronan {Le Bras}, Jianfeng Gao, and Yejin Choi.
  2020.
\newblock \href {https://aaai.org/ojs/index.php/AAAI/article/view/6239}
  {{PIQA:} reasoning about physical commonsense in natural language}.
\newblock In \emph{Proc. of {AAAI}}.

\bibitem[{Bosselut et~al.(2021)Bosselut, {Le Bras}, and Choi}]{Bosselut:2021}
Antoine Bosselut, Ronan {Le Bras}, and Yejin Choi. 2021.
\newblock \href {https://arxiv.org/abs/1911.03876} {Dynamic neuro-symbolic
  knowledge graph construction for zero-shot commonsense question answering}.
\newblock In \emph{Proc. of AAAI}.

\bibitem[{Bowman(2022)}]{Bowman:2021}
Samuel~R. Bowman. 2022.
\newblock \href {https://arxiv.org/abs/2110.08300} {The dangers of
  underclaiming: Reasons for caution when reporting how nlp systems fail}.
\newblock In \emph{Proc. of ACL}.

\bibitem[{Bowman et~al.(2015)Bowman, Angeli, Potts, and
  Manning}]{bowman2015large}
Samuel~R. Bowman, Gabor Angeli, Christopher Potts, and Christopher~D. Manning.
  2015.
\newblock \href {https://doi.org/10.18653/v1/D15-1075} {A large annotated
  corpus for learning natural language inference}.
\newblock In \emph{Proc. of EMNLP}.

\bibitem[{Brown et~al.(2020)Brown, Mann, Ryder, Subbiah, Kaplan, Dhariwal,
  Neelakantan, Shyam, Sastry, Askell, Agarwal, Herbert{-}Voss, Krueger,
  Henighan, Child, Ramesh, Ziegler, Wu, Winter, Hesse, Chen, Sigler, Litwin,
  Gray, Chess, Clark, Berner, McCandlish, Radford, Sutskever, and
  Amodei}]{Brown:2020}
Tom~B. Brown, Benjamin Mann, Nick Ryder, Melanie Subbiah, Jared Kaplan,
  Prafulla Dhariwal, Arvind Neelakantan, Pranav Shyam, Girish Sastry, Amanda
  Askell, Sandhini Agarwal, Ariel Herbert{-}Voss, Gretchen Krueger, Tom
  Henighan, Rewon Child, Aditya Ramesh, Daniel~M. Ziegler, Jeffrey Wu, Clemens
  Winter, Christopher Hesse, Mark Chen, Eric Sigler, Mateusz Litwin, Scott
  Gray, Benjamin Chess, Jack Clark, Christopher Berner, Sam McCandlish, Alec
  Radford, Ilya Sutskever, and Dario Amodei. 2020.
\newblock \href
  {https://proceedings.neurips.cc/paper/2020/hash/1457c0d6bfcb4967418bfb8ac142f64a-Abstract.html}
  {Language models are few-shot learners}.
\newblock In \emph{Proc. of NeurIPS}.

\bibitem[{Card et~al.(2020)Card, Henderson, Khandelwal, Jia, Mahowald, and
  Jurafsky}]{Card:2020}
Dallas Card, Peter Henderson, Urvashi Khandelwal, Robin Jia, Kyle Mahowald, and
  Dan Jurafsky. 2020.
\newblock \href {https://doi.org/10.18653/v1/2020.emnlp-main.745} {With little
  power comes great responsibility}.
\newblock In \emph{Proc. of EMNLP}.

\bibitem[{Card and Smith(2018)}]{card-smith-2018-importance}
Dallas Card and Noah~A. Smith. 2018.
\newblock \href {https://doi.org/10.18653/v1/N18-1148} {The importance of
  calibration for estimating proportions from annotations}.
\newblock In \emph{Proc. of NAACL}.

\bibitem[{Chang et~al.(2021)Chang, He, Jia, and
  Singh}]{chang-etal-2021-robustness}
Kai-Wei Chang, He~He, Robin Jia, and Sameer Singh. 2021.
\newblock \href {https://aclanthology.org/2021.emnlp-tutorials.5} {Robustness
  and adversarial examples in natural language processing}.
\newblock In \emph{Proc. of EMNLP: Tutorial Abstracts}.

\bibitem[{Chen et~al.(2020)Chen, Xu, Cheng, Xiaochuan, Zhang, Song, Wang, Qi,
  and Chu}]{chen-etal-2020-question}
Kunlong Chen, Weidi Xu, Xingyi Cheng, Zou Xiaochuan, Yuyu Zhang, Le~Song,
  Taifeng Wang, Yuan Qi, and Wei Chu. 2020.
\newblock \href {https://doi.org/10.18653/v1/2020.emnlp-main.549} {Question
  directed graph attention network for numerical reasoning over text}.
\newblock In \emph{Proc. of EMNLP}.

\bibitem[{Chow(1957)}]{Chow1957AnOC}
Chi-Keung Chow. 1957.
\newblock An optimum character recognition system using decision functions.
\newblock \emph{IRE Transactions on Electronic Computers}, 6:247--254.

\bibitem[{Cortes et~al.(2016)Cortes, DeSalvo, and Mohri}]{Cortes:2016}
Corinna Cortes, Giulia DeSalvo, and Mehryar Mohri. 2016.
\newblock \href
  {https://proceedings.neurips.cc/paper/2016/file/7634ea65a4e6d9041cfd3f7de18e334a-Paper.pdf}
  {Boosting with abstention}.
\newblock In \emph{Proc. of NeurIPS}.

\bibitem[{Davidov et~al.(2010)Davidov, Tsur, and
  Rappoport}]{davidov-etal-2010-semi}
Dmitry Davidov, Oren Tsur, and Ari Rappoport. 2010.
\newblock \href {https://aclanthology.org/W10-2914} {Semi-supervised
  recognition of sarcasm in {T}witter and {A}mazon}.
\newblock In \emph{Proc. of CoNLL}.

\bibitem[{DeGroot and Fienberg(1983)}]{Degroot:1983}
Morris~H. DeGroot and Stephen~E. Fienberg. 1983.
\newblock The comparison and evaluation of forecasters.
\newblock \emph{Journal of the Royal Statistical Society: Series D (The
  Statistician)}, 32(1-2):12--22.

\bibitem[{Devlin et~al.(2019)Devlin, Chang, Lee, and Toutanova}]{Devlin:2019}
Jacob Devlin, Ming-Wei Chang, Kenton Lee, and Kristina Toutanova. 2019.
\newblock \href {https://doi.org/10.18653/v1/N19-1423} {{BERT}: Pre-training of
  deep bidirectional transformers for language understanding}.
\newblock In \emph{Proc. of NAACL-HLT}.

\bibitem[{Dodge et~al.(2021)Dodge, Sap, Marasović, Agnew, Ilharco, Groeneveld,
  Mitchell, and Gardner}]{dodge2021documenting}
Jesse Dodge, Maarten Sap, Ana Marasović, William Agnew, Gabriel Ilharco, Dirk
  Groeneveld, Margaret Mitchell, and Matt Gardner. 2021.
\newblock \href {https://aclanthology.org/2021.emnlp-main.98} {Documenting
  large webtext corpora: A case study on the colossal clean crawled corpus}.
\newblock In \emph{Proc. of EMNLP}.

\bibitem[{Dua et~al.(2019)Dua, Wang, Dasigi, Stanovsky, Singh, and
  Gardner}]{dua-etal-2019-drop}
Dheeru Dua, Yizhong Wang, Pradeep Dasigi, Gabriel Stanovsky, Sameer Singh, and
  Matt Gardner. 2019.
\newblock \href {https://doi.org/10.18653/v1/N19-1246} {{DROP}: A reading
  comprehension benchmark requiring discrete reasoning over paragraphs}.
\newblock In \emph{Proc. of NAACL-HLT}.

\bibitem[{Eisenstein(2022)}]{Eisenstein:2022}
Jacob Eisenstein. 2022.
\newblock \href {https://doi.org/10.48550/ARXIV.2204.04487} {Uninformative
  input features and counterfactual invariance: Two perspectives on spurious
  correlations in natural language}.
\newblock In \emph{Proc. of NAACL}.

\bibitem[{Elazar et~al.(2021)Elazar, Zhang, Goldberg, and
  Roth}]{elazar-etal-2021-back}
Yanai Elazar, Hongming Zhang, Yoav Goldberg, and Dan Roth. 2021.
\newblock \href {https://aclanthology.org/2021.emnlp-main.819} {Back to square
  one: Artifact detection, training and commonsense disentanglement in the
  {W}inograd schema}.
\newblock In \emph{Proc. of EMNLP}.

\bibitem[{Gardner et~al.(2021)Gardner, Merrill, Dodge, Peters, Ross, Singh, and
  Smith}]{Gardner:2021}
Matt Gardner, William Merrill, Jesse Dodge, Matthew~E. Peters, Alexis Ross,
  Sameer Singh, and Noah~A. Smith. 2021.
\newblock \href {https://arxiv.org/abs/2104.08646} {Competency problems: On
  finding and removing artifacts in language data}.
\newblock In \emph{Proc. of EMNLP}.

\bibitem[{Gehman et~al.(2020)Gehman, Gururangan, Sap, Choi, and
  Smith}]{gehman2020realtoxicityprompts}
Samuel Gehman, Suchin Gururangan, Maarten Sap, Yejin Choi, and Noah~A. Smith.
  2020.
\newblock \href {https://doi.org/10.18653/v1/2020.findings-emnlp.301}
  {{R}eal{T}oxicity{P}rompts: Evaluating neural toxic degeneration in language
  models}.
\newblock In \emph{Findings of EMNLP}.

\bibitem[{Glockner et~al.(2018)Glockner, Shwartz, and
  Goldberg}]{glockner_acl18}
Max Glockner, Vered Shwartz, and Yoav Goldberg. 2018.
\newblock \href {https://doi.org/10.18653/v1/P18-2103} {Breaking {NLI} systems
  with sentences that require simple lexical inferences}.
\newblock In \emph{Proc. of ACL}.

\bibitem[{Goyal et~al.(2017)Goyal, Khot, Summers{-}Stay, Batra, and
  Parikh}]{Goyal:2017}
Yash Goyal, Tejas Khot, Douglas Summers{-}Stay, Dhruv Batra, and Devi Parikh.
  2017.
\newblock \href {https://doi.org/10.1109/CVPR.2017.670} {Making the {V} in
  {VQA} matter: Elevating the role of image understanding in visual question
  answering}.
\newblock In \emph{Proc. of CVPR}.

\bibitem[{Graesser(2013)}]{graesser2013prose}
Arthur~C Graesser. 2013.
\newblock \emph{Prose comprehension beyond the word}.

\bibitem[{Grice(1975)}]{grice1975logic}
Herbert~P Grice. 1975.
\newblock Logic and conversation.
\newblock In \emph{Speech acts}.

\bibitem[{Grice(1989)}]{grice1989studies}
Paul Grice. 1989.
\newblock \emph{Studies in the Way of Words}.

\bibitem[{Gu et~al.(2022)Gu, Han, Liu, and Huang}]{gu2021ppt}
Yuxian Gu, Xu~Han, Zhiyuan Liu, and Minlie Huang. 2022.
\newblock {PPT}: Pre-trained prompt tuning for few-shot learning.
\newblock In \emph{Proc. of ACL}.

\bibitem[{Guo et~al.(2017)Guo, Pleiss, Sun, and Weinberger}]{Guo:2017}
Chuan Guo, Geoff Pleiss, Yu~Sun, and Kilian~Q. Weinberger. 2017.
\newblock On calibration of modern neural networks.
\newblock In \emph{Proc. of ICML}.

\bibitem[{Gururangan et~al.(2018)Gururangan, Swayamdipta, Levy, Schwartz,
  Bowman, and Smith}]{Gururangan:2018}
Suchin Gururangan, Swabha Swayamdipta, Omer Levy, Roy Schwartz, Samuel Bowman,
  and Noah~A. Smith. 2018.
\newblock \href {https://doi.org/10.18653/v1/N18-2017} {Annotation artifacts in
  natural language inference data}.
\newblock In \emph{Proc. of NAACL-HLT}.

\bibitem[{He et~al.(2021{\natexlab{a}})He, Gao, and Chen}]{He:2021b}
Pengcheng He, Jianfeng Gao, and Weizhu Chen. 2021{\natexlab{a}}.
\newblock \href {https://doi.org/10.48550/ARXIV.2111.09543} {{DeBERTaV3}:
  Improving {DeBERTa} using {ELECTRA}-style pre-training with
  gradient-disentangled embedding sharing}.
\newblock {arXiv}:2111.09543.

\bibitem[{He et~al.(2021{\natexlab{b}})He, Liu, Gao, and Chen}]{He:2021a}
Pengcheng He, Xiaodong Liu, Jianfeng Gao, and Weizhu Chen. 2021{\natexlab{b}}.
\newblock \href {https://doi.org/10.48550/ARXIV.2006.03654} {{DeBERTa}:
  Decoding-enhanced {BERT} with disentangled attention}.
\newblock In \emph{Proc. of ICLR}.

\bibitem[{Hellman(1970)}]{Hellman1970TheNN}
Martin~E. Hellman. 1970.
\newblock The nearest neighbor classification rule with a reject option.
\newblock \emph{IEEE Transactions on Systems Science and Cybernetics},
  6:179--185.

\bibitem[{Hossain et~al.(2020)Hossain, Anastasopoulos, Blanco, and
  Palmer}]{hossain-etal-2020-non}
Md~Mosharaf Hossain, Antonios Anastasopoulos, Eduardo Blanco, and Alexis
  Palmer. 2020.
\newblock \href {https://doi.org/10.18653/v1/2020.findings-emnlp.345} {It{'}s
  not a non-issue: Negation as a source of error in machine translation}.
\newblock In \emph{Findings of EMNLP}.

\bibitem[{Hossain et~al.(2022)Hossain, Chinnappa, and Blanco}]{Hossain:2022}
Md~Mosharaf Hossain, Dhivya Chinnappa, and Eduardo Blanco. 2022.
\newblock \href {https://doi.org/10.48550/ARXIV.2203.08929} {An analysis of
  negation in natural language understanding corpora}.
\newblock In \emph{Proc. of ACL}.

\bibitem[{Hudson and Manning(2019)}]{hudson2019gqa}
Drew~A. Hudson and Christopher~D. Manning. 2019.
\newblock \href {https://doi.org/10.1109/CVPR.2019.00686} {{GQA:} {A} new
  dataset for real-world visual reasoning and compositional question
  answering}.
\newblock In \emph{Proc. of CVPR}.

\bibitem[{Kamath et~al.(2020)Kamath, Jia, and
  Liang}]{kamath-etal-2020-selective}
Amita Kamath, Robin Jia, and Percy Liang. 2020.
\newblock \href {https://doi.org/10.18653/v1/2020.acl-main.503} {Selective
  question answering under domain shift}.
\newblock In \emph{Proc. of ACL}.

\bibitem[{Laidlaw and Feizi(2019)}]{Laidlaw:2019}
Cassidy Laidlaw and Soheil Feizi. 2019.
\newblock \href {https://arxiv.org/abs/1911.11253} {Playing it safe:
  Adversarial robustness with an abstain option}.
\newblock {arXiv}:1911.11253.

\bibitem[{{Le Bras} et~al.(2020){Le Bras}, Swayamdipta, Bhagavatula, Zellers,
  Peters, Sabharwal, and Choi}]{bras2020adversarial}
Ronan {Le Bras}, Swabha Swayamdipta, Chandra Bhagavatula, Rowan Zellers,
  Matthew~E. Peters, Ashish Sabharwal, and Yejin Choi. 2020.
\newblock \href {http://proceedings.mlr.press/v119/bras20a.html} {Adversarial
  filters of dataset biases}.
\newblock In \emph{Proc. of ICML}.

\bibitem[{Lee et~al.(2022)Lee, Ippolito, Nystrom, Zhang, Eck, Callison-Burch,
  and Carlini}]{lee2021deduplicating}
Katherine Lee, Daphne Ippolito, Andrew Nystrom, Chiyuan Zhang, Douglas Eck,
  Chris Callison-Burch, and Nicholas Carlini. 2022.
\newblock Deduplicating training data makes language models better.
\newblock In \emph{Proc. of ACL}.

\bibitem[{Levesque et~al.(2012)Levesque, Davis, and
  Morgenstern}]{Levesque:2012}
Hector Levesque, Ernest Davis, and Leora Morgenstern. 2012.
\newblock The winograd schema challenge.
\newblock In \emph{Proc. of KR}.

\bibitem[{Li et~al.(2021)Li, Kuncoro, d'Autume, Blunsom, and
  Nematzadeh}]{Li:2021}
Xiang~Lorraine Li, Adhiguna Kuncoro, Cyprien de~Masson d'Autume, Phil Blunsom,
  and Aida Nematzadeh. 2021.
\newblock \href {https://doi.org/10.48550/ARXIV.2111.00607} {Do language models
  learn commonsense knowledge?}
\newblock {arXiv}:2111.00607.

\bibitem[{Liu and Singh(2004)}]{Liu2004}
Hugo Liu and Push Singh. 2004.
\newblock \href {http://citeseer.ist.psu.edu/liu04conceptnet.html} {Conceptnet:
  A practical commonsense reasoning toolkit}.
\newblock \emph{BT Technology Journal}, 22(4).

\bibitem[{Liu et~al.(2021{\natexlab{a}})Liu, Lee, Jia, and
  Liang}]{liu-lee-jia-liang:2021:arxiv}
Nelson~F. Liu, Tony Lee, Robin Jia, and Percy Liang. 2021{\natexlab{a}}.
\newblock \href {https://arxiv.org/abs/2102.01065} {Can small and synthetic
  benchmarks drive modeling innovation? a retrospective study of question
  answering modeling approaches}.
\newblock {arXiv}:2102.01065.

\bibitem[{Liu et~al.(2021{\natexlab{b}})Liu, Yuan, Fu, Jiang, Hayashi, and
  Neubig}]{Liu:2021}
Pengfei Liu, Weizhe Yuan, Jinlan Fu, Zhengbao Jiang, Hiroaki Hayashi, and
  Graham Neubig. 2021{\natexlab{b}}.
\newblock \href {https://arxiv.org/abs/2107.13586} {Pre-train, prompt, and
  predict: A systematic survey of prompting methods in natural language
  processing}.
\newblock {arXiv}:2107.13586.

\bibitem[{Liu et~al.(2019)Liu, Ott, Goyal, Du, Joshi, Chen, Levy, Lewis,
  Zettlemoyer, and Stoyanov}]{liu2019roberta}
Yinhan Liu, Myle Ott, Naman Goyal, Jingfei Du, Mandar Joshi, Danqi Chen, Omer
  Levy, Mike Lewis, Luke Zettlemoyer, and Veselin Stoyanov. 2019.
\newblock \href {https://arxiv.org/abs/1907.11692} {{RoBERTa}: A robustly
  optimized bert pretraining approach}.
\newblock {arXiv}:1907.11692.

\bibitem[{Merrill et~al.(2021)Merrill, Goldberg, Schwartz, and
  Smith}]{Merrill:2021}
Will Merrill, Yoav Goldberg, Roy Schwartz, and Noah~A. Smith. 2021.
\newblock \href {https://arxiv.org/abs/2104.10809} {Provable limitations of
  acquiring meaning from ungrounded form:what will future language models
  understand?}
\newblock \emph{TACL}.

\bibitem[{Mohammad et~al.(2016)Mohammad, Shutova, and
  Turney}]{mohammad-etal-2016-metaphor}
Saif Mohammad, Ekaterina Shutova, and Peter Turney. 2016.
\newblock \href {https://doi.org/10.18653/v1/S16-2003} {Metaphor as a medium
  for emotion: An empirical study}.
\newblock In \emph{Proc. of *{SEM}}.

\bibitem[{Morante and Blanco(2012)}]{morante-blanco-2012-sem}
Roser Morante and Eduardo Blanco. 2012.
\newblock \href {https://aclanthology.org/S12-1035} {*{SEM} 2012 shared task:
  Resolving the scope and focus of negation}.
\newblock In \emph{Proc. of *{SEM}}.

\bibitem[{Oprea and Magdy(2020)}]{Oprea:2020}
Silviu Oprea and Walid Magdy. 2020.
\newblock \href {https://doi.org/10.18653/v1/2020.acl-main.118} {i{S}arcasm: A
  dataset of intended sarcasm}.
\newblock In \emph{Proc. of ACL}.

\bibitem[{Pearl(2009)}]{pearl2009causality}
Judea Pearl. 2009.
\newblock \emph{Causality}.
\newblock Cambridge university press.

\bibitem[{Peters et~al.(2018)Peters, Neumann, Iyyer, Gardner, Clark, Lee, and
  Zettlemoyer}]{Peters:2018}
Matthew~E. Peters, Mark Neumann, Mohit Iyyer, Matt Gardner, Christopher Clark,
  Kenton Lee, and Luke Zettlemoyer. 2018.
\newblock \href {https://doi.org/10.18653/v1/N18-1202} {Deep contextualized
  word representations}.
\newblock In \emph{Proc. of NAACL-HLT}.

\bibitem[{Peters et~al.(2019)Peters, Neumann, Logan, Schwartz, Joshi, Singh,
  and Smith}]{Peters:2019}
Matthew~E. Peters, Mark Neumann, Robert Logan, Roy Schwartz, Vidur Joshi,
  Sameer Singh, and Noah~A. Smith. 2019.
\newblock \href {https://doi.org/10.18653/v1/D19-1005} {Knowledge enhanced
  contextual word representations}.
\newblock In \emph{Proc. of EMNLP}.

\bibitem[{Poliak et~al.(2018)Poliak, Naradowsky, Haldar, Rudinger, and
  Van~Durme}]{poliak-etal-2018-hypothesis}
Adam Poliak, Jason Naradowsky, Aparajita Haldar, Rachel Rudinger, and Benjamin
  Van~Durme. 2018.
\newblock \href {https://doi.org/10.18653/v1/S18-2023} {Hypothesis only
  baselines in natural language inference}.
\newblock In \emph{Proc. of *{SEM}}.

\bibitem[{Qin et~al.(2020)Qin, Shwartz, West, Bhagavatula, Hwang, {Le Bras},
  Bosselut, and Choi}]{Qin:2020}
Lianhui Qin, Vered Shwartz, Peter West, Chandra Bhagavatula, Jena~D. Hwang,
  Ronan {Le Bras}, Antoine Bosselut, and Yejin Choi. 2020.
\newblock \href {https://doi.org/10.18653/v1/2020.emnlp-main.58} {Back to the
  future: Unsupervised backprop-based decoding for counterfactual and abductive
  commonsense reasoning}.
\newblock In \emph{Proc. of EMNLP}.

\bibitem[{Radford et~al.(2019)Radford, Wu, Child, Luan, Amodei, Sutskever
  et~al.}]{radford2019language}
Alec Radford, Jeffrey Wu, Rewon Child, David Luan, Dario Amodei, Ilya
  Sutskever, et~al. 2019.
\newblock Language models are unsupervised multitask learners.
\newblock \emph{OpenAI blog}, 1(8).

\bibitem[{Raffel et~al.(2020)Raffel, Shazeer, Roberts, Lee, Narang, Matena,
  Zhou, Li, and Liu}]{Raffel:2020}
Colin Raffel, Noam Shazeer, Adam Roberts, Katherine Lee, Sharan Narang, Michael
  Matena, Yanqi Zhou, Wei Li, and Peter~J. Liu. 2020.
\newblock \href {http://jmlr.org/papers/v21/20-074.html} {Exploring the limits
  of transfer learning with a unified text-to-text transformer}.
\newblock \emph{JMLR}, 21(140):1--67.

\bibitem[{Raji et~al.(2021)Raji, Bender, Paullada, Denton, and
  Hanna}]{raji2021ai}
Inioluwa~Deborah Raji, Emily~M. Bender, Amandalynne Paullada, Emily Denton, and
  Alex Hanna. 2021.
\newblock \href {https://arxiv.org/abs/2111.15366} {{AI} and the everything in
  the whole wide world benchmark}.
\newblock In \emph{Proc. Of NeurIPS Benchmarks and Datasets track}.

\bibitem[{Rajpurkar et~al.(2018)Rajpurkar, Jia, and Liang}]{Rajpurkar:2018}
Pranav Rajpurkar, Robin Jia, and Percy Liang. 2018.
\newblock \href {https://doi.org/10.18653/v1/P18-2124} {Know what you don{'}t
  know: Unanswerable questions for {SQ}u{AD}}.
\newblock In \emph{Proc. of ACL}.

\bibitem[{Rajpurkar et~al.(2016)Rajpurkar, Zhang, Lopyrev, and
  Liang}]{rajpurkar2016squad}
Pranav Rajpurkar, Jian Zhang, Konstantin Lopyrev, and Percy Liang. 2016.
\newblock \href {https://doi.org/10.18653/v1/D16-1264} {{SQ}u{AD}: 100,000+
  questions for machine comprehension of text}.
\newblock In \emph{Proc. of EMNLP}.

\bibitem[{Ray et~al.(2016)Ray, Christie, Bansal, Batra, and Parikh}]{Ray:2016}
Arijit Ray, Gordon Christie, Mohit Bansal, Dhruv Batra, and Devi Parikh. 2016.
\newblock \href {https://doi.org/10.18653/v1/D16-1090} {Question relevance in
  {VQA}: Identifying non-visual and false-premise questions}.
\newblock In \emph{Proc. of EMNLP}.

\bibitem[{Roberts et~al.(2020)Roberts, Raffel, and
  Shazeer}]{roberts2020knowledge}
Adam Roberts, Colin Raffel, and Noam Shazeer. 2020.
\newblock \href {https://doi.org/10.18653/v1/2020.emnlp-main.437} {How much
  knowledge can you pack into the parameters of a language model?}
\newblock In \emph{Proc. of EMNLP}.

\bibitem[{Rogers(2021)}]{rogers-2021-changing}
Anna Rogers. 2021.
\newblock \href {https://doi.org/10.18653/v1/2021.acl-long.170} {Changing the
  world by changing the data}.
\newblock In \emph{Proc. of ACL}.

\bibitem[{Rubinstein et~al.(2015)Rubinstein, Levi, Schwartz, and
  Rappoport}]{Rubinstein:2015}
Dana Rubinstein, Effi Levi, Roy Schwartz, and Ari Rappoport. 2015.
\newblock \href {https://doi.org/10.3115/v1/P15-2119} {How well do
  distributional models capture different types of semantic knowledge?}
\newblock In \emph{Proc. of ACL}.

\bibitem[{Sakaguchi et~al.(2020)Sakaguchi, {Le Bras}, Bhagavatula, and
  Choi}]{sakaguchi2020winogrande}
Keisuke Sakaguchi, Ronan {Le Bras}, Chandra Bhagavatula, and Yejin Choi. 2020.
\newblock \href {https://arxiv.org/abs/1907.10641} {{WinoGrande}: An
  adversarial winograd schema challenge at scale}.
\newblock In \emph{Proc. of AAAI}.

\bibitem[{Schick and Sch{\"u}tze(2021)}]{schick2021its}
Timo Schick and Hinrich Sch{\"u}tze. 2021.
\newblock \href {https://doi.org/10.18653/v1/2021.naacl-main.185} {It{'}s not
  just size that matters: Small language models are also few-shot learners}.
\newblock In \emph{Proc. of NAACL}.

\bibitem[{Schick and Schütze(2021)}]{Schick:2021}
Timo Schick and Hinrich Schütze. 2021.
\newblock \href {https://arxiv.org/abs/2111.13440} {True few-shot learning with
  prompts -- a real-world perspective}.
\newblock {arXiv}:2111.13440.

\bibitem[{Schwartz et~al.(2020)Schwartz, Dodge, Smith, and
  Etzioni}]{Schwartz:2020}
Roy Schwartz, Jesse Dodge, Noah~A. Smith, and Oren Etzioni. 2020.
\newblock \href {https://doi.org/10.1145/3381831} {Green {AI}}.
\newblock \emph{CACM}, 63(12).

\bibitem[{Schwartz et~al.(2017)Schwartz, Sap, Konstas, Zilles, Choi, and
  Smith}]{Schwartz:2017}
Roy Schwartz, Maarten Sap, Ioannis Konstas, Leila Zilles, Yejin Choi, and
  Noah~A. Smith. 2017.
\newblock \href {https://doi.org/10.18653/v1/K17-1004} {The effect of different
  writing tasks on linguistic style: A case study of the {ROC} story cloze
  task}.
\newblock In \emph{Proc. of CoNLL}.

\bibitem[{Sharma et~al.(2018)Sharma, Allen, Bakhshandeh, and
  Mostafazadeh}]{sharma-etal-2018-tackling}
Rishi Sharma, James Allen, Omid Bakhshandeh, and Nasrin Mostafazadeh. 2018.
\newblock \href {https://doi.org/10.18653/v1/P18-2119} {Tackling the story
  ending biases in the story cloze test}.
\newblock In \emph{Proc. of ACL}.

\bibitem[{Shin et~al.(2020)Shin, Razeghi, Logan~IV, Wallace, and
  Singh}]{shin2020autoprompt}
Taylor Shin, Yasaman Razeghi, Robert~L. Logan~IV, Eric Wallace, and Sameer
  Singh. 2020.
\newblock \href {https://doi.org/10.18653/v1/2020.emnlp-main.346}
  {{A}uto{P}rompt: {E}liciting {K}nowledge from {L}anguage {M}odels with
  {A}utomatically {G}enerated {P}rompts}.
\newblock In \emph{Proc. of EMNLP}.

\bibitem[{Shrikumar et~al.(2019)Shrikumar, Alexandari, and
  Kundaje}]{shrikumar2019flexible}
Avanti Shrikumar, Amr Alexandari, and Anshul Kundaje. 2019.
\newblock \href {https://arxiv.org/abs/1802.07024} {A flexible and adaptive
  framework for abstention under class imbalance}.
\newblock {arXiv}:1802.07024.

\bibitem[{Simonyan et~al.(2014)Simonyan, Vedaldi, and
  Zisserman}]{simonyan2014deep}
Karen Simonyan, Andrea Vedaldi, and Andrew Zisserman. 2014.
\newblock \href {http://arxiv.org/abs/1312.6034} {Deep inside convolutional
  networks: Visualising image classification models and saliency maps}.
\newblock {arXiv}:1312.6034.

\bibitem[{Smilkov et~al.(2017)Smilkov, Thorat, Kim, Viégas, and
  Wattenberg}]{smilkov2017smoothgrad}
Daniel Smilkov, Nikhil Thorat, Been Kim, Fernanda Viégas, and Martin
  Wattenberg. 2017.
\newblock \href {https://arxiv.org/abs/1706.03825} {{SmoothGrad}: removing
  noise by adding noise}.
\newblock {arXiv}:1706.03825.

\bibitem[{Socher et~al.(2013)Socher, Perelygin, Wu, Chuang, Manning, Ng, and
  Potts}]{socher-etal-2013-recursive}
Richard Socher, Alex Perelygin, Jean Wu, Jason Chuang, Christopher~D. Manning,
  Andrew Ng, and Christopher Potts. 2013.
\newblock \href {https://aclanthology.org/D13-1170} {Recursive deep models for
  semantic compositionality over a sentiment treebank}.
\newblock In \emph{Proc. of EMNLP}.

\bibitem[{Stanovsky et~al.(2019)Stanovsky, Smith, and
  Zettlemoyer}]{Stanovsky:2019}
Gabriel Stanovsky, Noah~A. Smith, and Luke Zettlemoyer. 2019.
\newblock \href {https://doi.org/10.18653/v1/P19-1164} {Evaluating gender bias
  in machine translation}.
\newblock In \emph{Proc. of ACL}.

\bibitem[{Steedman and Baldridge(2006)}]{ccg}
M.~Steedman and J.~Baldridge. 2006.
\newblock \href
  {https://doi.org/https://doi.org/10.1016/B0-08-044854-2/02028-9} {Combinatory
  categorial grammar}.
\newblock In Keith Brown, editor, \emph{Encyclopedia of Language \& Linguistics
  (Second Edition)}, second edition edition, pages 610--621. Elsevier, Oxford.

\bibitem[{Strubell et~al.(2019)Strubell, Ganesh, and McCallum}]{Strubell:2019}
Emma Strubell, Ananya Ganesh, and Andrew McCallum. 2019.
\newblock \href {https://doi.org/10.18653/v1/P19-1355} {Energy and policy
  considerations for deep learning in {NLP}}.
\newblock In \emph{Proc. of ACL}.

\bibitem[{Sulem et~al.(2021)Sulem, Hay, and Roth}]{sulem-etal-2021-know-dont}
Elior Sulem, Jamaal Hay, and Dan Roth. 2021.
\newblock \href {https://aclanthology.org/2021.findings-emnlp.385} {Do we know
  what we don{'}t know? studying unanswerable questions beyond {SQ}u{AD} 2.0}.
\newblock In \emph{Findings of EMNLP}.

\bibitem[{Sundararajan et~al.(2017)Sundararajan, Taly, and
  Yan}]{sundararajan2017axiomatic}
Mukund Sundararajan, Ankur Taly, and Qiqi Yan. 2017.
\newblock \href {http://proceedings.mlr.press/v70/sundararajan17a.html}
  {Axiomatic attribution for deep networks}.
\newblock In \emph{Proc. of ICML}.

\bibitem[{Swayamdipta et~al.(2020)Swayamdipta, Schwartz, Lourie, Wang,
  Hajishirzi, Smith, and Choi}]{Swayamdipta:2020}
Swabha Swayamdipta, Roy Schwartz, Nicholas Lourie, Yizhong Wang, Hannaneh
  Hajishirzi, Noah~A. Smith, and Yejin Choi. 2020.
\newblock \href {https://doi.org/10.18653/v1/2020.emnlp-main.746} {Dataset
  cartography: Mapping and diagnosing datasets with training dynamics}.
\newblock In \emph{Proc. of EMNLP}.

\bibitem[{Tandon et~al.(2020)Tandon, Sakaguchi, Dalvi, Rajagopal, Clark,
  Guerquin, Richardson, and Hovy}]{tandon-etal-2020-dataset}
Niket Tandon, Keisuke Sakaguchi, Bhavana Dalvi, Dheeraj Rajagopal, Peter Clark,
  Michal Guerquin, Kyle Richardson, and Eduard Hovy. 2020.
\newblock \href {https://doi.org/10.18653/v1/2020.emnlp-main.520} {A dataset
  for tracking entities in open domain procedural text}.
\newblock In \emph{Proc. of EMNLP}.

\bibitem[{Tsvetkov et~al.(2014)Tsvetkov, Boytsov, Gershman, Nyberg, and
  Dyer}]{tsvetkov-etal-2014-metaphor}
Yulia Tsvetkov, Leonid Boytsov, Anatole Gershman, Eric Nyberg, and Chris Dyer.
  2014.
\newblock \href {https://doi.org/10.3115/v1/P14-1024} {Metaphor detection with
  cross-lingual model transfer}.
\newblock In \emph{Proc. of ACL}.

\bibitem[{Wang et~al.(2021)Wang, Li, Aslan, and Vinyals}]{Wang:2021}
Luyu Wang, Yujia Li, Ozlem Aslan, and Oriol Vinyals. 2021.
\newblock \href {https://aclanthology.org/2021.textgraphs-1.7} {{W}iki{G}raphs:
  A {W}ikipedia text - knowledge graph paired dataset}.
\newblock In \emph{Proc. of TextGraphs}.

\bibitem[{Wang and Culotta(2020)}]{wang-culotta-2020-identifying}
Zhao Wang and Aron Culotta. 2020.
\newblock \href {https://doi.org/10.18653/v1/2020.findings-emnlp.308}
  {Identifying spurious correlations for robust text classification}.
\newblock In \emph{Findings of EMNLP}.

\bibitem[{Weller and Seppi(2019)}]{weller2019humor}
Orion Weller and Kevin Seppi. 2019.
\newblock \href {https://doi.org/10.18653/v1/D19-1372} {Humor detection: A
  transformer gets the last laugh}.
\newblock In \emph{Proc. of EMNLP}.

\bibitem[{Yaghoobzadeh et~al.(2021)Yaghoobzadeh, Mehri, Tachet~des Combes,
  Hazen, and Sordoni}]{yaghoobzadeh-etal-2021-increasing}
Yadollah Yaghoobzadeh, Soroush Mehri, Remi Tachet~des Combes, T.~J. Hazen, and
  Alessandro Sordoni. 2021.
\newblock \href {https://doi.org/10.18653/v1/2021.eacl-main.291} {Increasing
  robustness to spurious correlations using forgettable examples}.
\newblock In \emph{Proc. of EACL}.

\bibitem[{Zellers et~al.(2018)Zellers, Bisk, Schwartz, and
  Choi}]{zellers-etal-2018-swag}
Rowan Zellers, Yonatan Bisk, Roy Schwartz, and Yejin Choi. 2018.
\newblock \href {https://doi.org/10.18653/v1/D18-1009} {{SWAG}: A large-scale
  adversarial dataset for grounded commonsense inference}.
\newblock In \emph{Proc. of EMNLP}.

\bibitem[{Zellers et~al.(2019)Zellers, Holtzman, Bisk, Farhadi, and
  Choi}]{zellers-etal-2019-hellaswag}
Rowan Zellers, Ari Holtzman, Yonatan Bisk, Ali Farhadi, and Yejin Choi. 2019.
\newblock \href {https://doi.org/10.18653/v1/P19-1472} {{H}ella{S}wag: Can a
  machine really finish your sentence?}
\newblock In \emph{Proc. of ACL}.

\bibitem[{Zhang et~al.(2018)Zhang, Liu, Liu, Gao, Duh, and
  Van~Durme}]{Zhang:2018}
Sheng Zhang, Xiaodong Liu, Jingjing Liu, Jianfeng Gao, Kevin Duh, and Benjamin
  Van~Durme. 2018.
\newblock \href {https://doi.org/10.48550/ARXIV.1810.12885} {{ReCoRD}: Bridging
  the gap between human and machine commonsense reading comprehension}.
\newblock {arXiv}:1810.12885.

\bibitem[{Zhang et~al.(2021)Zhang, Gong, and Choi}]{zhang-etal-2021-knowing}
Shujian Zhang, Chengyue Gong, and Eunsol Choi. 2021.
\newblock \href {https://doi.org/10.18653/v1/2021.findings-acl.172} {Knowing
  more about questions can help: Improving calibration in question answering}.
\newblock In \emph{Findings of ACL}.

\bibitem[{Zhang et~al.(2019)Zhang, Han, Liu, Jiang, Sun, and
  Liu}]{zhang-etal-2019-ernie}
Zhengyan Zhang, Xu~Han, Zhiyuan Liu, Xin Jiang, Maosong Sun, and Qun Liu. 2019.
\newblock \href {https://doi.org/10.18653/v1/P19-1139} {{ERNIE}: Enhanced
  language representation with informative entities}.
\newblock In \emph{Proc. of ACL}.

\bibitem[{Zhao et~al.(2017)Zhao, Wang, Yatskar, Ordonez, and Chang}]{Zhao:2017}
Jieyu Zhao, Tianlu Wang, Mark Yatskar, Vicente Ordonez, and Kai-Wei Chang.
  2017.
\newblock \href {https://doi.org/10.18653/v1/D17-1323} {Men also like shopping:
  Reducing gender bias amplification using corpus-level constraints}.
\newblock In \emph{Proc. of EMNLP}.

\end{thebibliography}
\bibliographystyle{acl_natbib}




\end{document}